%% file: 00_Article_Merge.tex
\documentclass[11pt,a4paper]{article}
\usepackage[hyperref]{acl2019}
\usepackage{times}
\usepackage{latexsym}
\usepackage{todonotes}
\usepackage{amsthm,amsmath,amssymb}
\usepackage[utf8]{inputenc}
\usepackage[T1]{fontenc}
\usepackage{booktabs}

\usepackage{url}

\aclfinalcopy 

\title{
Semantic Change and Emerging Tropes \\ In a Large Corpus of New High German Poetry
}

\author{Thomas Nikolaus Haider \\
  MPI for Empirical Aesthetics, Frankfurt\\
  IMS, University of Stuttgart \\
  \texttt{thomas.haider@ae.mpg.de} \\\And
  Steffen Eger \\
  Natural Language Learning Group \\
  Technical University of Darmstadt \\
  \texttt{eger@aiphes.tu-darmstadt.de} \\}

\date{}

\begin{document}
\maketitle

\begin{abstract}

Due to its semantic succinctness and novelty of expression, poetry is a great test bed for semantic change analysis. However, so far there is a scarcity of large diachronic corpora. Here, we provide a large corpus of German poetry which consists of about 75k poems with more than 11 million tokens, with poems ranging from the 16th to early 20th century. We then track semantic change in this corpus by investigating the rise of tropes (`love is magic') over time and detecting change points of meaning, which we find to occur particularly within the German Romantic period. Additionally, through self-similarity, we reconstruct literary periods and find evidence that the law of linear semantic change also applies to poetry.


\end {abstract}
    \input{01_Article_Introduction}

    \input{02_Article_RelatedWork}

\input{03_Article_Method}
    \input{04_Article_Corpus}

    \input{05_Article_Results}

    \input{06_Article_Conclusion}

\bibliographystyle{acl_natbib}
\bibliography{07_Bibliography_Clean}

\end{document}

%% file: 01_Article_Introduction.tex


\section{Introduction}
Following in the footsteps of traditional poetry analysis, Natural Language Understanding (NLU) research has largely explored \textit{stylistic variation} \citep{kaplan2007computational,kao2015computational}, (over time) \citep{voigt2013tradition}, with a focus on \textit{sound devices} \citep{mccurdy2015poemage,hench2017phonological} and broadly canonised form features such as \textit{meter} \citep{greene2010automatic,agirrezabal-etal-2016-machine,estes2016supervised} and \textit{rhyme} \citep{reddy2011unsupervised,haider2018supervised}, 
as well as 
\textit{enjambement} \citep{ruiz2017enjambment} and \textit{noun+noun metaphor} \citep{kesarwani2017metaphor}.

However, poetry also lends itself well to semantic change analysis, as linguistic invention \citep{underwood2012,herbelot2014semantics} and succinctness \citep{roberts2000poetry} are at the core of poetic production. Poetic language is generally very dense, where concepts / ideas cannot be easily paraphrased. With a distributional semantics model, \citet{herbelot2014semantics} finds that the coherence of poetry significantly differs from Wikipedia and random text, allowing the conclusion that poetry is -- compared to ordinary language -- unusual in its word choice, but still generally regarded comprehensible language. 
Recently, there has been research with topic models on poetry with Latent Dirichlet Allocation. \citet{navarro2018poetic} explores the overarching topical motifs in a corpus of Spanish sonnetts, while \citet{haider2019diachronic} sketches the evolution of topics over time in a German poetry corpus, identifying salient topics for certain literature periods and applying these for downstream learning how to date a poem.

Following in this vein, we offer a method to explore poetic tropes, i.e.\ word pairs such as `love (is) magic' that gain association strength (cosine similarity) over time, finding that most are gaining traction in the Romantic period. Further, we track the self-similarity of words, both with a change point analysis and by evaluating `total self-similarity' of words over time. The former helps us to reconstruct literary periods, while the latter provides us with further evidence for the law of linearity of semantic change \citep{eger-mehler-2016-linearity} 
using our 
new 
method.

We do this with a model that learns diachronic word2vec embeddings jointly over all our time slots \citep{bamman2014distributed}, avoiding the need to compute the cosine similarity of two word vector representations on second order to align the embeddings.

Our contributions are: we (1) provide a large corpus of German poetry which consists of about 75k poems, ranging from the 16th to early 20th century with more than 11 million tokens.\footnote{\url{http://github.com/thomasnikolaushaider/DLK}}
We then track semantic change in this corpus with (2) two self-similarity experiments and finally (3)  by investigating the rise of tropes (e.g. `love is magic') over time. 



 

%% file: 02_Article_RelatedWork.tex
\section{Related Work}
Semantic change has been explored in various works in recent years. 
One focus has been on studying laws of semantic change. 
\citet{Xu:2015} explore two earlier proposed laws quantitatively: the law of differentiation (near-synonyms tend to differentiate over time) and the law of parallel change (related words have analogous meaning changes), finding that the latter applies more broadly. 
\citet{hamilton-etal-2016-diachronic} find that frequent words  have a lower chance of undergoing semantic change and more polysemous words are more likely to change semantically. 
\citet{eger-mehler-2016-linearity} find that semantic change is linear in two senses: semantic self-similarity of words tends to decrease linearly in time and word vectors at time $t$ can be written as linear combinations of words vectors at time $t-1$, which allows to forecast meaning change. Regarding methods, \citet{Xu:2015} work with simple distributional count vectors, while \citet{hamilton-etal-2016-diachronic} and \citet{eger-mehler-2016-linearity} use low-dimensional dense vector representations. Both works use different approaches to map independently induced word vectors (across time) in a common space: \citet{hamilton-etal-2016-diachronic} learn to align word vectors using a projection matrix while \citet{eger-mehler-2016-linearity} induce second-order embeddings by computing the similarity of words, in each time slot, to a reference vocabulary. \citet{kutuzov-etal-2018-diachronic} survey and compare models of semantic change based on diachronic word embeddings. \citet{dubossarsky2017} caution against confounds in semantic change models.

An interesting approach besides computing independent word embeddings in each time period has been outlined by \citet{bamman2014distributed} who \emph{jointly} compute embeddings across different linguistic variables: each word $w$ has an embedding 
\begin{align*}
\mathbf{w}=\mathbf{e}_w\mathbf{W}_{\text{main}}+\mathbf{e}_w\mathbf{W}_C,
\end{align*}
where $\mathbf{W}_{\text{main}}\in \mathbb{R}^{|V|\times d}$ is a main embedding matrix and $\mathbf{W}_C\in\mathbb{R}^{|V|\times d}$ is an embedding matrix for linguistic variable $C$, and $\mathbf{e}_w$ is a 1-hot vector (index) of word $w$. In their original work, $C$ ranges over geographic locations (US states). A joint model has several advantages: it better addresses data sparsity (for specific variables) and it directly learns to map words in a joint vector space without necessity of ex-post  projection. In our work, we use this latter model for temporal embeddings in that each linguistic variable $C$ corresponds to a time epoch $t$:
\begin{align*}
    \mathbf{w}(t) = \mathbf{e}_w\mathbf{W}_{\text{main}}+\mathbf{e}_w\mathbf{W}_t
\end{align*}

This dispenses the need to align independently trained embeddings for every time slot. Instead, a joint (MAIN) model is learned that is then re-weighted for every time epoch. While this is convenient, it does not necessarily mean that embeddings of a certain low-frequency word in a given time slot are stable. If there is not enough context for a given word in a certain time period $t$, the model just learns the MAIN embedding with little to no re-weighting, i.e., the matrix $\mathbf{W}_t$ may not be well estimated (at certain rows).

%% file: 04_Article_Corpus.tex
\section*{Corpus}
We compile the largest corpus of poetry to date, the \textbf{German Poetry Corpus v1}, or Deutsches Lyrik Korpus version 1, \textbf{DLK} for short. See table \ref{tab:corpus} for a size overview. We know of no larger poetry collections in any language. Only the collection from the English Project Gutenberg offers a similar size, but due to a lawsuit, as of 2018 it is not available in Germany anymore.\footnote{\url{http://block.pglaf.org/germany.shtml}}



 \begin{table}[ht]
 \center 
\begin{tabular}{r|r}
\toprule
Tokens & 11,849,112  \\
Lines & 1,784,613  \\
Stanzas & 280,234  \\
Poems & 74,155  \\
Authors & 269 \\
\bottomrule
\end{tabular}
\caption{Corpus Size, Deutsches Lyrik Korpus v1}
    \label{tab:corpus}
\end{table}

DLK covers the full range of the New High German language (of public domain literature), ranging from 1575 AD (Barock period) up to 1936 AD (Modern period). It is collected from three resources: (1) Textgrid\footnote{\url{textgrid.de}} (TGRID), (2) The German Text Archive\footnote{\url{deutschestextarchiv.de}} (DTA), and (3) Antikoerperchen (ANTI-K). The latter two were first described by \citet{haider2018supervised}. All three corpora are set in TEI P5 XML.

TGRID offers around 51k poems with the label `verse' (TGRID-V). Many of these texts have a unique timestamp. Where this is not the case, we take the average year between the author's birth and death.

DTA offers around 28k poems with the label `lyrik' (DTA-L). The poetry in DTA is organized by editions (whole books), rather than by single poems. The timestamps are therefore guided by these few books, but give very accurate stamps.

ANTI-K is a collection of only 156 poems of school canon that was mined from \url{antikoerperchen.de/lyrik}. It has very accurate annotation, including literary periods, that allow us to gauge the distribution of poems according to canonic periods.

For training our model, we organize the corpus by stanzas, where every stanza represents a document. The reasoning behind this is that for poetic tropes, words are likely to stand in local context. We merge our collections and remove duplicate stanzas that match on their first line. This removes 9600 duplicates. Filtering Dutch and French material further eliminates 3200 stanzas. Since the earliest time slot 1575--1625 is too small, we merge it with the adjacent slot.

 \begin{figure}[ht]
    \centering
    \includegraphics[width=1\linewidth]{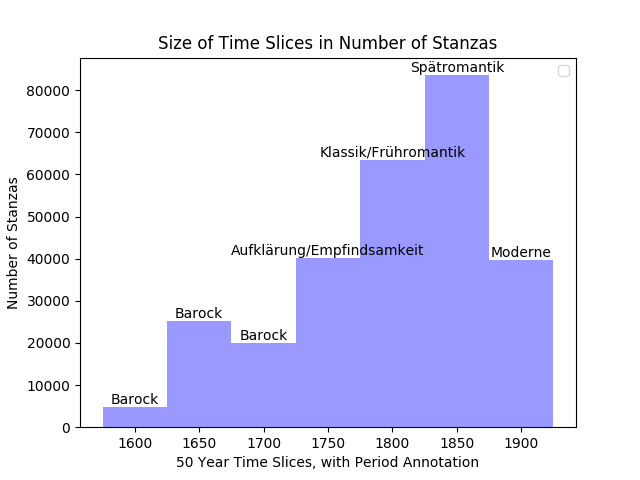}
    \caption{Distribution of stanzas in 50 year slots, 1575--1925 AD. Period labels: Barock (baroque), Aufklärung (enlightenment), Empfindsamkeit (sentimentalism), Klassik (Weimar classicism), Frühromantik (early romantic), Spätromantik (late romantic), Moderne (modernity). First slot (1600) is merged into the adjacent slot.}
    \label{fig:corpus}
 \end{figure}

See figure \ref{fig:corpus} for the distribution of stanzas in 50 year time slots. The slots are labelled with approximate literature period information based on the clustered annotation in ANTI-K. We can see that the Romantic period (approx.\ 1750--1875) is overly heavy, while the Barock period is somewhat underrepresented.

We lemmatize based on a gold token:lemma mapping that was extracted from DTA-L in tcf format. Where this does not cover a token, we pos-tag the line with \textit{pattern.de} to feed into germalemma.\footnote{\url{https://github.com/WZBSocialScienceCenter/germalemma}} We publish our corpus in json format.\footnote{\url{http://github.com/thomasnikolaushaider/DLK}}


%% file: 05_Article_Results.tex
\section*{Experiments}

\subsection*{Self-similarity}

We investigate semantic self-similarity of words over time in two ways: (1) How does poetic diction change over successive time steps (change point detection), and (2) how does contextual word meaning change in total over the whole time frame with respect to the word's frequency (laws of conformity and linearity)?
 We use a model with a 25+50 sliding window, where time steps increase by 25 years, with a window size of 50 years. This doubles the data and allows a more fine grained analysis.

\subsubsection*{Pairwise Self-Similarity}

We compute how the contextual use of words changes over successive time steps. We do this by determining the self-similarity of a word $w$ over time by calculating the cosine 
similarity 
of the embedding vectors $\mathbf{w}(t)$ for $w$ at time periods $t=t_i$ and $t=t_{i+1}$ as in 
equation (\ref{eq:cos}):

\begin{align}\label{eq:cos}
    \text{cossim}(\mathbf{w}(t_i), \mathbf{w}(t_{i+1}))
\end{align}
where cossim$(\mathbf{a},\mathbf{b})$ is defined as $\mathbf{a}^\intercal\mathbf{b}$ for two normalized vectors $\mathbf{a}$ and $\mathbf{b}$.

\begin{figure}[!ht]
    \centering
    \includegraphics[width=1\linewidth]{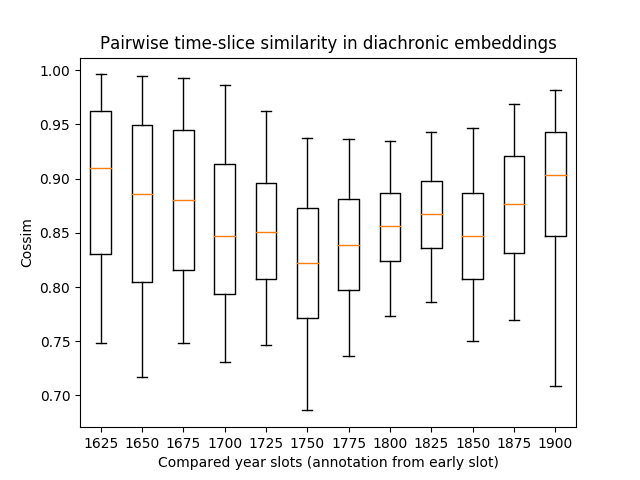}
    \caption{\textbf{Pairwise Self-Similarity. Top-3000 most frequent words. Cossine similarities of word w with itself in adjacent time slots $cossim(w(t_i),w(t_{i+1}))$}}
    \label{fig:pairwise}
 \end{figure}

Thus, we can aggregate the self-similarity for the most frequent words at every time step and plot the change for all these words combined. See figure \ref{fig:pairwise} for a boxplot of this pairwise self-similarity for the 3000 most frequent words.

\subsubsection*{Results}

 Our interpretation 
 is 
 that rising similarity signifies a homogenization of overall word use (diction), while a falling similarity signifies semantic diversification. In particular, we see a steady falling trajectory in the period between 1600 and 1675, with a dip at 1700. This period is generally regarded as the `Barock' period. Then, word use slowly homogenizes, until we see a sharp dip around 1750, the onset of the Romantic period. Then it homogenizes during the Romantic period, until a dip at 1850, the end of the Romantic period, and then a homogenization into the the onset of modernity.



 \subsubsection*{Total Self-Similarity}
 
 We determine 
 change of word meaning across any possible time distances as a probing for the linearity of semantic change in our corpus. 
 
For this, we calculate the semantic self-similarity of a word  
across all time periods $t_i$ and $t_j$ with $t_i<t_j$. 
We then aggregate all pairwise distances in years 
\begin{align*}
    \text{dist}(t_i,t_j)= |t_i-t_j|
\end{align*} for all words  $w$.\footnote{For all 25,50,\ldots,300 year distances, 
cossims per word in these distances are averaged, so we are left with one value per distance and word.} To obtain robust estimates of embeddings, we only allow words that occur at least 50 times in every time slot and remove stopwords, leaving us with 472 words.

 \begin{figure}[!ht]
    \centering
    \includegraphics[width=1\linewidth]{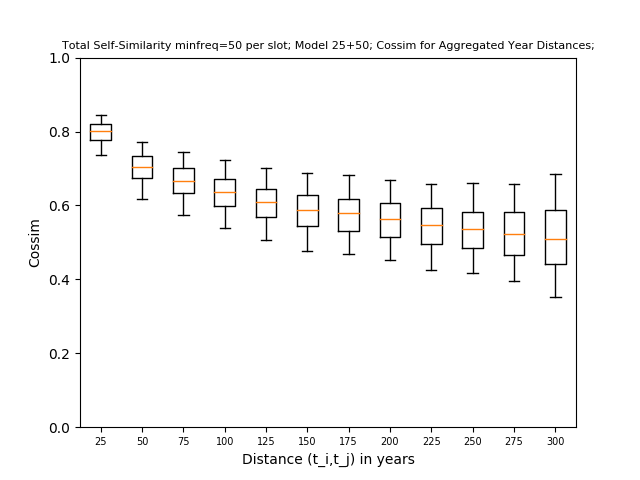}
    \caption{\textbf{Total Self-Similarity of words that occur at least 50 times in every time slot. Cossine similarities aggregated by the distance of compared time slots $(t_i,t_j)$ averaged for every time slot given a word. Removed stopwords. Whiskers: [5,95] percentiles.}}
    \label{fig:totalsimlinear}
 \end{figure}

 The x-axis in Figure \ref{fig:totalsimlinear} gives the distances $\text{dist}(t_i,t_j)$ while the y-axis shows the distribution of cossims over all words $w$ within each distance.

 We find that there is approximately a  linear relation between the distance of timeslots for an average word, where close slots are more similar, and far apart slots are increasingly dissimilar. However, the variance also increases with distance.
 
 Additionally, we equally divide our words into a low-frequency and a high-frequency band. We find that the low-frequency band shows a generally higher self-similarity than the high-frequency band over all distances. This would mean that, overall, high frequency words tend to be more semantically diverse over time, i.e.\ stand in more diverse contexts. In contrast, low-frequency words stand in fewer contexts, therefore undergo less change. However, this could also come from the tendency of the model to revert to MAIN. 

\subsection*{Emerging Tropes: Collocations \& Metaphors}
\subsubsection*{Method}
To detect emerging tropes, we calculate the cosine similarity of word pairs over time. For the sake of visualization we use a 50+50 model with 6 time slots. We then perform Principal Component Analysis (PCA) over the resulting trajectories \cite{Eger:2010}. The resulting principal components show that similar trajectories are co-variant. Component 1 aggregates stable high/low trajectories, while component 2 aggregates rising/falling trajectories. We illustrate our finding with the tropes for the concept `love' (`Liebe' in German) and determine the most salient word pairs over the whole dataset. `Love' is a very frequent word in poetry. Nevertheless, this approach works equally well for any word, except very low frequency words that show idiosyncratic behavior as they are not well distributed.

\subsubsection*{Results}
We calculate the distance of `love' against every other word $w$, where $w$ has to occur at least 30 times in the corpus, and it needs to be represented in every time slot at least twice. We allow one slot to be empty. 

The first 4 components of PCA explain >.95 variance, where component 1 explains 73\%, component 2 13\%, and component 3 5\%. 
We retrieve the top-25 word pairs at every component extreme.

 \begin{table}[ht]
 \begin{small}
\begin{tabular}{r|r|r|r}
 \textbf{rising traj.} & \textbf{falling traj.} & \textbf{high traj.} & \textbf{low traj.}  \\
 \hline
 \hline
frische         & aufrechen     & liebe         & brummen   \\
veilchen        & alsbald       & freundschaft  & krähen   \\
niedersinken    & billigkeit    & lust          & rasseln  \\
duftig          & erzeigen      & treue         & rum   \\
jenseits        & unterstehen   & trieb         & bock   \\
zauber          & betragen      & seligkeit     & dum  \\
entgleiten      & stracks       & hoffnung      & prasseln  \\
künden          & zuerkennen    & glaube        & trommel   \\
hoffend         & hierin        & keusch        & säbel \\
efeu            & schmeissen    & treu          & traben \\
enthüllen       & anlaß         & erkalten      & bellen    \\
erfüllung       & jederzeit     & wahr          & block \\
heimat          & muhen         & immerdar      & bügel \\
trübe           & schimpfen     & regung        & gaul  \\
gloria          & stecken       & gegenliebe    & grasen \\
\end{tabular}
\end{small}
\caption{Top 15 words per dimension for 'love' tropes from PCA extremes, plotted in figures 4, 5, 6 and 7.}
\label{tab:wordtable}
\end{table}

\begin{figure}[!ht]
\begin{minipage}[b]{0.48\linewidth}
\centering
\includegraphics[width=\textwidth]{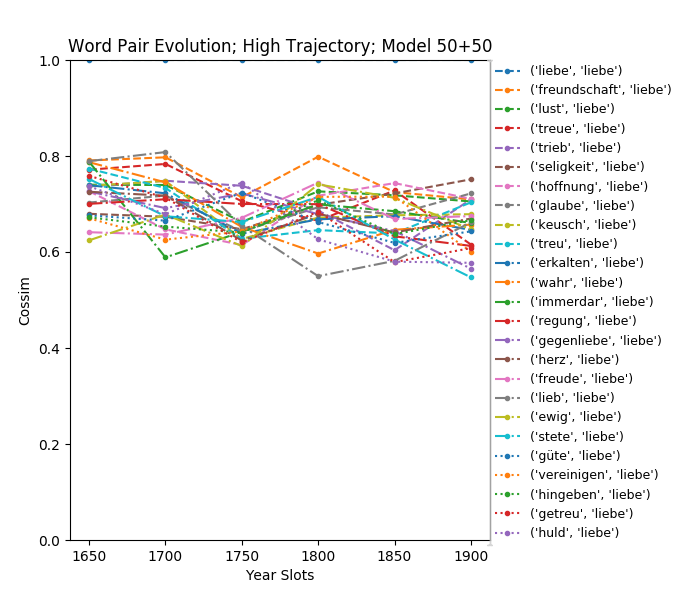}
\caption{Love: High}
\label{fig:figure1}
\end{minipage}
\hspace{0.001cm}
\begin{minipage}[b]{0.48\linewidth}
\centering
\includegraphics[width=\textwidth]{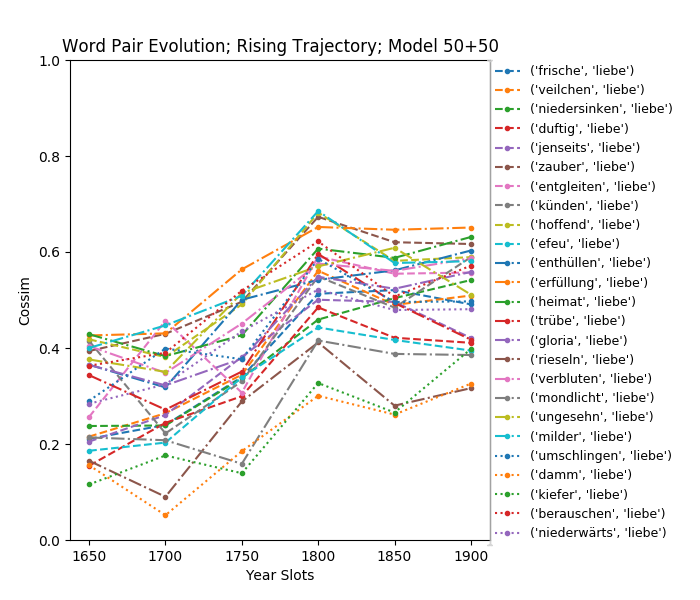}
\caption{Love: Rising}
\label{fig:figure2}
\end{minipage}
\end{figure}

\begin{figure}[!ht]
\begin{minipage}[b]{0.48\linewidth}
\centering
\includegraphics[width=\textwidth]{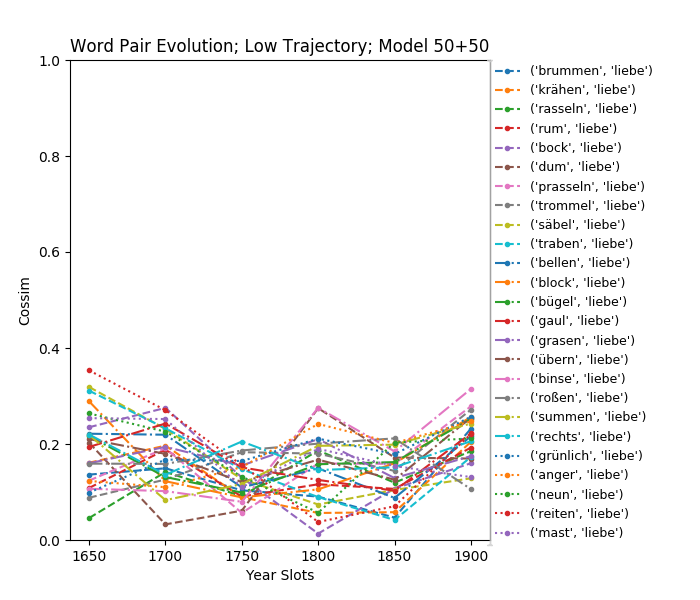}
\caption{Love: Low}
\label{fig:figure3}
\end{minipage}
\hspace{0.001cm}
\begin{minipage}[b]{0.48\linewidth}
\centering
\includegraphics[width=\textwidth]{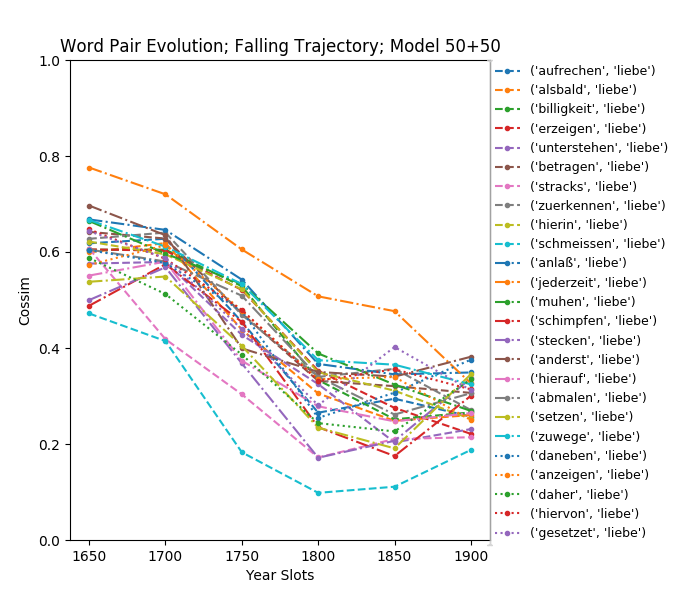}
\caption{Love: Falling}
\label{fig:figure4}
\end{minipage}
\end{figure}

We find that component 1 orders trajectories based on 
high/low semantic similarity, while component 2 orders based on rising/falling trajectories. See figures \ref{fig:figure1} (high trajectory), \ref{fig:figure2} (rising trajectory), \ref{fig:figure3} (low trajectory) and \ref{fig:figure4} (falling trajectory). 
See table \ref{tab:wordtable} for the respective word pairs (collocations) with `love' as they are plotted.

  \paragraph{\textbf{Stable High Trajectories}}
 Trajectories in figure \ref{fig:figure1} (table \ref{tab:wordtable} column 3) have a consistently high cosine, meaning that these collocations have remained unchanged since the Baroque period: `love is fidelity',\footnote{(`Treue', `Liebe')} `love is friendship',\footnote{('Freundschaft', 'Liebe')} or `love is lust'. These are conventional near-synonyms, just as (`apple', `tree')\footnote{(`Apfel', `Baum')} or idioms (`apples', `pears').\footnote{(`Äpfel', `Birnen'), 'compare apples and oranges'.}
 A k-nearest neighbor (KNN) analysis retrieves these collocations. Performing this analysis for multiple words, we find that the idiom (`apple', `pear') is a special case, as it strongly loads into both rising and stable high PCA dimensions (both top 20). 
 
 \paragraph{\textbf{Rising Trajectories}}
 Figure \ref{fig:figure2} (table \ref{tab:wordtable} column 1) shows rising collocations that emerge during the Romantic period, i.e.\ `fresh love',\footnote{('Frische', 'Liebe')} `love is magic / enchantment'\footnote{(`Zauber', `Liebe')} and 'love is violets'.\footnote{(`Veilchen', 'Liebe')} A metaphorical (trope) interpretation is most likely here.
 

  \paragraph{\textbf{Falling Trajectories}} As illustrated in figure \ref{fig:figure4} (column 2), these  collocations fall into obscurity. We find `cheap love'\footnote{billigkeit (which changed from 'equity' to 'cheap')} or things like `raking'\footnote{aufrechen} or `manners / accounting'.\footnote{betragen}
 
  \paragraph{\textbf{Stable Low Trajectories}}
 The lines in figure \ref{fig:figure3} (column 4) signify word pairs that are always far apart. We find things that make noise, like drums.\footnote{trommel} The `drums of love' seems to be an oxymoron.

%% file: 06_Article_Conclusion.tex
\section*{Conclusion}
We constructed the largest poetry corpus to date and investigated distributional semantic change with different methods. With self-similarity, we can reconstruct literature period transitions and find that the the law of linear semantic change also applies to poetry. However, for confident analysis of other laws more data and a more robust model is still called for. Finally, we extract emerging and vanishing poetic tropes based on the co-variance of time trajectories of word pairs. This method is applicable more broadly to cluster similar trajectories for any given word pairs. We found trajectories of word similarities that are beyond simple nearest-neighbor analysis, and illustrated findings for reasonable tropes with 'love'. While large, our dataset is still somewhat sparse in the distribution of words over all time slots, partially because many word forms simply emerge / vanish at a certain point ('excitement' is not in Baroque).






